\pdfoutput=1

\documentclass[11pt]{article}

\usepackage{acl}

\usepackage{times}
\usepackage{latexsym}
\usepackage{xspace}
\usepackage{tabularx}
\usepackage{booktabs}
\usepackage{multirow}
\newcolumntype{Y}{>{\centering\arraybackslash}X}
\newcolumntype{L}{>{\raggedright\arraybackslash}X}
\def\gD{{\mathcal{T}}}
\usepackage{algorithm}
\usepackage[noend]{algpseudocode}
\usepackage{amsmath,amsfonts,bm}
\usepackage{graphicx}
\graphicspath{ {./images/} }

\usepackage[T1]{fontenc}

\usepackage[utf8]{inputenc}

\usepackage{microtype}

\usepackage{xspace}
\usepackage{tabularx}
\newcommand{\ttbf}[1]{\texttt{\textbf{#1}}}
\newcommand{\da}[0]{\ttbf{PromDA}\xspace}

\usepackage{amsmath,bm}
\usepackage{amsfonts}

\def\eqref#1{equation~\ref{#1}}

\def\1{\bm{1}}

\def\vh{{\bm{h}}}

\def\vp{{\bm{p}}}

\def\vw{{\bm{w}}}

\DeclareMathAlphabet{\mathsfit}{\encodingdefault}{\sfdefault}{m}{sl}
\SetMathAlphabet{\mathsfit}{bold}{\encodingdefault}{\sfdefault}{bx}{n}

\def\gD{{\mathcal{D}}}

\def\gT{{\mathcal{T}}}

\def\emP{{P}}

\title{\da: Prompt-based Data Augmentation for Low-Resource NLU Tasks}

\author{Yufei Wang$^{1}$\thanks{~~Work done during the internship at Microsoft STCA.}, Can Xu$^{2}$, Qingfeng Sun$^{2}$, Huang Hu$^{2}$, Chongyang Tao$^{2}$ \\ \textbf{Xiubo Geng}$^{2}$ \and \textbf{Daxin Jiang}$^{2}$\thanks{~~Corresponding author} \\
Macquarie University, Sydney, Australia$^1$ \\
Microsoft Corporation, Beijing, China$^2$ \\
\texttt{yufei.wang@students.mq.edu.au} \\
\texttt{\{caxu,qins,huahu,chongyang.tao,xigeng,djiang\}@microsoft.com} \\
}

\begin{document}
\maketitle
\begin{abstract}
This paper focuses on the \emph{Data Augmentation} for low-resource Natural Language Understanding (NLU) tasks. We propose \textbf{Prompt}-based \textbf{D}ata \textbf{A}ugmentation model (\da) which only trains small-scale \emph{Soft Prompt} (i.e., a set of trainable vectors) in the frozen Pre-trained Language Models (PLMs). This avoids human effort in collecting \emph{unlabeled in-domain data} and maintains the quality of generated synthetic data. In addition, \da 
generates synthetic data via two different views and filters out the low-quality data using NLU models. Experiments on four benchmarks show that synthetic data produced by \da successfully boost up the performance of NLU models which consistently outperform several competitive baseline models, including a state-of-the-art semi-supervised model using \emph{unlabeled in-domain data}. The synthetic data from \da are also complementary with \emph{unlabeled in-domain data}. The NLU models can be further improved when they are combined for training.
\end{abstract}

\section{Introduction}
Deep neural networks often require large-scale high-quality labeled training data to achieve state-of-the-art performance~\cite{bowman-etal-2015-large}. However, constructing labeled data could be challenging in many scenarios~\cite{feng-etal-2021-survey}. In this paper, we study the low-resource Natural Language Understanding (NLU) tasks, including sentence classification and sequence labelling tasks, where only small labeled data is available. Previous works often produce extra ``labeled data'' for the NLU models to learn. ~\citet{wang2021meta} deploys the \emph{self-training} framework to produce \emph{pseudo labelled training data} from \emph{unlabeled in-domain data} which could be expensive to obtain.~\citet{xu-etal-2021-augnlg} has shown that extracting domain-specific unlabeled data from the general corpus is not trivial. ~\citet{wei-zou-2019-eda,dai-adel-2020-analysis} expand the original small training data using automatic heuristic rules, such as randomly synonyms replacement, which effectively creates new training instances. However, these processes may distort the text, making the generated syntactic data grammatically and semantically incorrect. 

To solve the above dilemma, many existing works~\cite{ding-etal-2020-daga,yang-etal-2020-generative,notenoughpaper} resort to applying Language Models (LMs) or Pre-trained Language Models (PLMs) for data augmentation in a low-resource setting. Given the labeled data, one can directly fine-tune PLMs to generate new synthetic data without additional human effort. However, we argue that, in the low-resource NLU tasks, directly fine-tuning all parameters of PLMs with small training data (especially when there are less than 100 samples) could result in over-fitting and PLMs simply \emph{memorizes} the training instances. As a result, the generated synthetic data could be very similar to the original training instances and cannot provide new training signals to the NLU models. Recently, several works~\cite{lester2021power,li-liang-2021-prefix} propose prompt tuning, which only back-propagates the error to \emph{Soft Prompts} (i.e., a sequence of continuous vectors prepended to the input of PLMs) instead of the entire model. They show that prompt tuning is sufficient to be competitive with full model tuning while significantly reducing the amount of parameters to be tuned. Thus, the prompt tuning is quite suitable to tackle the above over-fitting issue in low-resource generative fine-tuning, which spawns more novel samples relative to the small labeled data under the premise of ensuring generation quality. 

Motivated by this, we propose \textbf{Prompt}-based \textbf{D}ata \textbf{A}ugmentation model (\da). Specifically, we freeze the entire pre-trained model and only allow tuning the additional soft prompts during fine-tuning on the small labeled training data. In addition, we have observed that the initialization of soft prompts has a significant impact on fine-tuning, especially when the low-resource situation reaches an extreme extent. To better initialize the prompt parameters for the data augmentation tasks, we propose \emph{task-agnostic} \emph{Synonym Keyword to Sentence} pre-training task to directly pre-train the prompt parameters of PLMs on their pre-training corpora. This task simulates the process of generating entire training sample from partial fragment information (e.g., keywords). Similar to previous works~\cite{ding-etal-2020-daga,yang-etal-2020-generative,notenoughpaper}, we could fine-tune PLMs to produce complete synthetic data conditioned on the output tags. We refer this as \emph{Output View Generation}. To boost the diversity of the generated samples, we introduce another fine-tuning generative task named \emph{Input View Generation}, which takes the extracted keywords from the sample as the input and the sample as the output. As NLG models trained from small training data still has a certain chance to generate low-quality samples, we leverage the NLU Consistency Filtering~\cite{notenoughpaper} to filter the generated samples.


We conduct experiments on four benchmarks: sequence labelling task CoNLL03~\cite{tjong-kim-sang-de-meulder-2003-introduction} and Wikiann~\cite{pan-etal-2017-cross}, sentence classification task SST-2~\cite{socher-etal-2013-recursive} and RT~\cite{pang-lee-2005-seeing}. Experiment results show that NLU models trained on synthetic data from \da consistently outperform several competitive baseline models, including a state-of-the-art semi-supervised NLU models MetaST~\citep{wang2021meta} on Sequence Labelling task. In addition, we find that the synthetic data from \da are also complementary with the \emph{unlabeled in-domain data}. The performance of NLU models can be further improved when both of them are combined. Finally, we conduct diversity analysis and case study to further confirm the synthetic data quality from \da. Our source code is released at~\url{https://github.com/GaryYufei/PromDA}.

\section{Related Work}
\paragraph{Prompt Learning}
The concept of prompt-based learning starts from the GPT3 model~\cite{NEURIPS2020_1457c0d6}. Previous works design different prompts to query language models to extract knowledge triples~\cite{petroni-etal-2019-language} or classify sentences into pre-defined categories~\cite{schick-schutze-2021-exploiting} in the few-shot setting. They construct various discrete prompts manually for these tasks. To reduce the human effort in this selection process, ~\cite{gao-etal-2021-making} proposes to expand prompts using pre-trained language models. However, the selection of discrete prompts is still an independent process and difficult to be optimized together with the downstream tasks in an end-to-end manner. ~\citet{DBLP:journals/corr/abs-2102-12206} proposes a complicated two-stage model to connect between prompt generation and downstream tasks. To solve this issue, ~\cite{lester2021power,li-liang-2021-prefix} propose to use soft prompts, which are sets of trainable vectors, in the frozen pre-trained language models. Unlike the hard prompts, these vectors do not correspond to any real words. It allows the optimization with the downstream tasks in an end-to-end manner. As shown in~\citet{li-liang-2021-prefix}, PLMs with \emph{Soft Prompts} can often perform better in the low-resource setting.

\paragraph{Generative Data Augmentation}
~\citet{hou-etal-2018-sequence} generates diverse utterances to improve dialogue understanding models.~\citet{xia-etal-2019-generalized} uses a bilingual dictionary and an unsupervised machine translation model to expand low-resource machine translation training data.~\citet{wu2019conditional,kumar-etal-2020-data} make use of the masking mechanism in many PLM pre-training objective functions (e.g., BERT~\cite{devlin-etal-2019-bert}, BART~\cite{lewis-etal-2020-bart}) and produce new synthetic data by masking randomly chosen words in the original training instances. ~\citet{ding-etal-2020-daga,yang-etal-2020-generative,notenoughpaper} apply LMs and PLMs to learn directly to generate new synthetic data for NLU tasks (i.e., sequence labeling and commonsense inference tasks after trained (fine-tuned) on the relatively large training data. These works often directly apply \emph{off-the-shelf} LMs or PLMs to generate synthetic data. ~\citet{DBLP:journals/corr/abs-2109-09193} proposes to use unlabelled data as \emph{hard prompt} to generate synthetic data without any training, limiting its application in complicated NLP tasks. To best of our knowledge, \da is the first PLMs with \emph{Soft Prompt} that are especially designed for the data augmentation task.

\begin{figure*}[!htb]
\includegraphics[width=\linewidth]{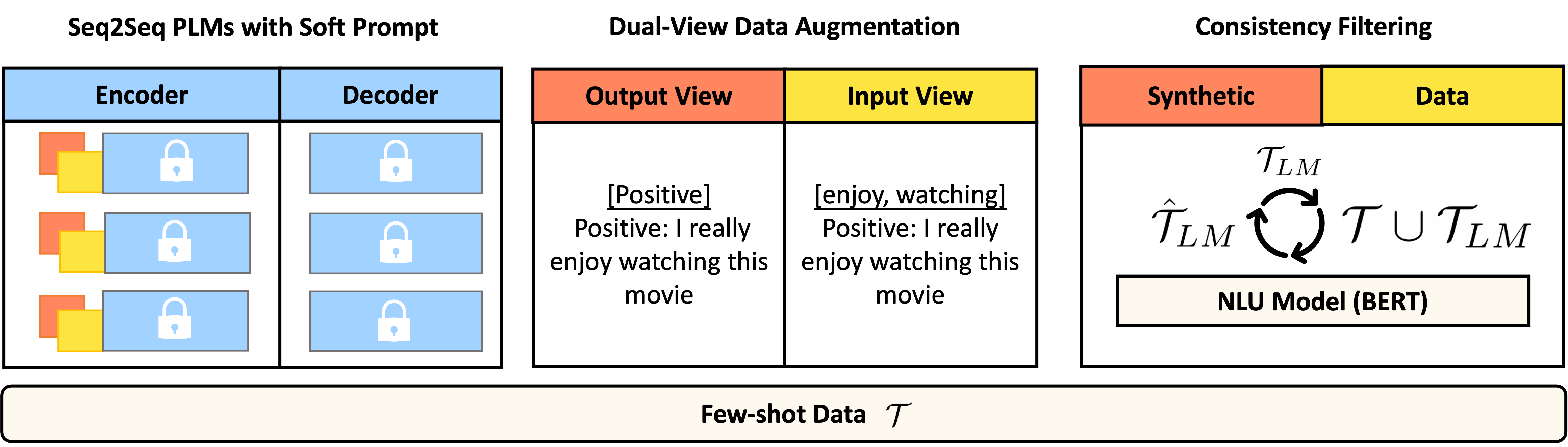}
\caption{The Overall of \da. \emph{Soft Prompt} prepend a sequence of trainable vector at each layer of the frozen PLMs. The white locker represents frozen parameters. We have separated sets of \emph{Soft Prompt} to support \emph{Daul-View Data Augmentation} where the \emph{Output View} conditions on the output tags and \emph{Input View} conditions on the keywords in the input sentences. Finally, we use the NLU models to iteratively filter out low-quality synthetic data and use the remaining synthetic data, combined with $\gT$, to train stronger NLU models.}
\label{overall}
\end{figure*}

\section{Prompt-based Data Augmentation}
This section first formulates the data augmentation for low-resource NLU task. We then introduce the three important components in Our proposed Prompt-based Data Augmentation method (\da), including \emph{i)} prompt-based learning in pre-trained language models; \emph{ii)} dual synthetic data generation view and \emph{iii)} \emph{Consistency Filtering}. Figure~\ref{overall} shows the overall of \da.

\subsection{Data Augmentation For NLU tasks}
In the low-resource NLU tasks, only a set of labeled training data $\gT=\{(x_1, y_1), \cdots, (x_n, y_n)\}$ is available where $n$ is relatively small (i.e., less than a hundred). \emph{Data Augmentation} generates synthetic labeled training data $\gT_{LM}=\{(\hat{x}_1, \hat{y}_1), \cdots, (\hat{x}_n, \hat{y}_n)\}$ from the original labeled training data $T$ using language models. The goal is that the NLU models trained using $\gT \cup \gT_{LM}$ outperform the NLU models only trained using $\gT$.

\subsection{Prompt-based learning}
Fine-tuning is the prevalent way to adapt PLMs to specific down-stream tasks~\cite{devlin-etal-2019-bert}. However, for low-resource data augmentation, we expect the generated synthetic training data $\gT_{LM}$ to be different from $\gT$ and to provide new information for NLU models to learn. A fine-tuned PLM, which is biased towards a small number of training instances, may not be an optimal solution.

Prompt-based learning, starting from the zero-shot instructions in GPT3~\cite{NEURIPS2020_1457c0d6}, keeps the whole PLMs parameters frozen and only prepends the discrete natural language task instructions (e.g. ``translate to English'') before the task inputs. Freezing the PLMs parameters might help generalization during training. However, finding suitable discrete task introductions cannot be easily optimized in an end-to-end fashion and requires extra human effort. In this paper, inspired by the recent work~\cite{lester2021power,li-liang-2021-prefix}, we replace the task introductions with \emph{Soft Prompt} (i.e., a sequence of continuous and trainable vectors). During training, we only update the parameters of this \emph{Soft Prompt} and fix all PLMs parameters. We mainly focus on generating synthetic training data using seq2seq Transformer-based PLMs.

Unlike~\citet{lester2021power} which only prepends \emph{Soft Prompt} at the input layer, inspired by Adaptor~\cite{houlsby2019parameter} which adds trainable Multi-layer Perceptron (MLP) at each transformer layer, we prepend a sequence of trainable vectors at each transformer layer. We denote $\emP^j=\{\vp^j_1, \cdots, \vp^j_k\}$ as the \emph{Soft Prompt} at the $j^{th}$ layer. The $i^{th}$ hidden states at the $j^{th}$ layer $\vh^j_i$ in the Transformer model is defined as follows:
\begin{equation}
\vh^j_i=\left\{
\begin{array}{ccl}
\vp^j_i & & {i \leq k}  \\
\vw_i & & {i > k \wedge j = 0} \\
\mathit{Trans}(\vh^{j-1})_i & & {\text{Otherwise}}
\end{array} \right.
\end{equation}
where $\mathit{Trans}(\dot)$ is the forward function the Transformer layer and $w_i$ is the fixed word embedding vector at the input layer. Compared to~\citep{lester2021power}, this allows gradients to be updated at each layer and better complete the learning tasks.

\subsection{Pre-training for Prompt Initialization}
The parameter initialization of the \emph{Soft Prompt} $\emP$ has a significant impact on the generated synthetic data quality, especially in the low-resource \emph{Data Augmentation} task.~\citet{lester2021power} proposes to further pre-train the full PLMs parameters, without the prompt parameters, to enhance the prompt capability. However, this strategy (i.e., full PLM pre-training) introduces significant computation overhead and does not provide any insight about prompt initialization. Instead, we propose to directly pre-train the parameters of the \emph{Soft Prompt} with the frozen PLMs. Given that data augmentation produces full syntactic data from partial information (e.g., output tags and keywords), we propose \emph{Synonym Keywords to Sentence} pre-training task. Given a chunk of text, we extract keywords using unsupervised keyword extraction algorithm \emph{Rake}~\cite{rose2010automatic}. We randomly replace some of these extracted keywords with their synonyms, via WordNet~\cite{fellbaum2010wordnet}. Given these synonym keywords, the \emph{Soft Prompt} is pre-trained to reconstruct the original text chunks. When applying this \emph{Soft Prompt} for data augmentation, we only need to fine-tune the \emph{Soft Prompt} with the few-shot labeled data $\gT$. This pre-training process only happens once. We only use the task-agnostic general-purpose pre-training corpus.

\begin{algorithm}[t!]
\caption{Dual-View Data Augmentation: Given few-shot labeled dataset $\gT$, the number of iteration $N$; return a trained NLU model $M_{NLU}$.}
\label{alg:data_diverse}
\begin{algorithmic}[1]
\Procedure{DualViewDA}{$\gD, N$}
    \State $M_{LM} \gets \textsc{Train}(LM, \gT)$
    \State $\gT_{I}^1 \gets \textsc{Gen}(M_{LM}, \gT, \textsc{I})$ \algorithmiccomment{Input}
    \State $\gT_{O}^1 \gets \textsc{Gen}(M_{LM}, \gT, \textsc{O})$ \algorithmiccomment{Output}
    \State $\gT_{I}^2 \gets \textsc{Gen}(M_{LM}, \gT_{O}^1, \textsc{I})$ 
    \State $\gT_{O}^2 \gets \textsc{Gen}(M_{LM}, \gT_{I}^1, \textsc{O})$
    \State $\hat{\gT}_{LM} \gets \gT_{I}^1 \cup \gT_{I}^2 \cup \gT_{O}^1 \cup \gT_{O}^2$
    \State $M_{NLU}^{0} \gets \textsc{Train}(NLU, \gT)$ 
    \For{$r \in 1, \ldots, N$}
        \State $\gT_{LM}^r \gets \textsc{Consist}(M_{NLU}^{r-1},\hat{\gT}_{LM})$
        \State $\gT^{r} \gets \gT_{LM}^r \cup \gT$
        \State $M_{NLU}^{r} \gets \textsc{Train}(NLU, \gT^{r})$
    \EndFor
    \State $M_{NLU} \gets M_{NLU}^{N}$
    \State \textbf{return} $M_{NLU}$
\EndProcedure
\end{algorithmic}
\end{algorithm}

\subsection{Dual-View Data Augmentation}
Previous works often restrict the encoder inputs to fixed keywords or limited labels, such as unconditional generation~\cite{yang-etal-2020-generative} and label-conditional generation~\cite{notenoughpaper}. The relatively small input space could result in similar outputs. To enrich the input space, we propose \emph{Dual-View} Data Augmentation that generates synthetic data from \emph{Input View}, which is conditioned on the keywords in the input sentences, and \emph{Output View}, which is conditioned on the output labels. Table~\ref{dualviewexample} shows examples of these two views. As illustrated in Algorithm~\ref{alg:data_diverse} (line 2 to 7), after fine-tuning the \emph{Soft Prompt} in PLMs, \da first generates $\gT_{I}^1$ and $\gT_{O}^1$ from \emph{Input View} and \emph{Output View}, respectively. \da then extracts output labels from $\gT_{I}^1$ and keywords from $\gT_{O}^1$. These new output labels and keywords are fed into the \emph{Output View} and \emph{Input View} in $M_{LM}$ to generate another two sets of new synthetic data $\gT_{O}^2$ and $\gT_{I}^2$. In this way, the resulting output text should maintain a higher level of diversity and include more novel words/phrases/knowledge.

\paragraph{Dual View via Prompt Ensemble}
Ensembles of different neural models can often achieve better performance~\cite{hansen1990neural}. Prompt-based learning provides an efficient way to model ensemble. By training $K$ sets of \emph{Soft Prompt}, we create $K$ models sharing the same frozen PLMs. In our case, after prompt pre-training, we treat \emph{Input View} and \emph{Output View} as two independent models and use the \emph{Soft Prompt} parameters $\emP$ to initialize the parameters of $\emP_{input}$ and $\emP_{output}$. During the \da fine-tuning, the gradients from the \emph{Input View} and \emph{Output View} training instances are only applied to parameters $\emP_{input}$ and $\emP_{output}$, respectively. This prompt ensemble allows the two views to generate synthetic data independently. As a result, the final output should include diverse real-world knowledge.

\subsection{Consistency Filtering}
As \da is trained from small training data, it is possible to generate low-quality samples. We leverage the NLU Consistency Filtering~\cite{notenoughpaper} to filter the generated samples. Specifically, given synthetic data with generated labels produced by \da, we use the NLU models to label these data again and only keep the instances with \emph{consistent} outputs from \da and the NLU models. As shown in Algorithm~\ref{alg:data_diverse} (line 8 to 12), $M_{NLU}^{r}$ filters the raw synthetic data $\hat{\gT}_{LM}$ into $\gT_{LM}$ which are combined with few-shot labeled data $\gT$ to train new NLU models $M_{NLU}^{r+1}$. As $M_{NLU}^{r+1}$ is generally better than $M_{NLU}^{r}$, we iterate this process $N$ times to obtain stronger NLU models.

\begin{table}[!ht]
\small
\centering
\begin{tabularx}{\linewidth}{cL}
\toprule
\multicolumn{2}{c}{{\bf Sequence Labelling}} \\
\midrule
GT: & \textbf{[\textit{Org} All Fishermen 's Association]} secretary \textbf{[\textit{Per} N.J. Bose]} said the strike would continue indefinitely. \\
IV: & \underline{All Fishermen 's Association} and \underline{N.J. Bose} and \underline{strike} and \underline{indefinitely} \\
OV: & \underline{Organization} and \underline{Person} \\
\midrule
\multicolumn{2}{c}{{\bf Sentence Classification}} \\
\midrule
GT: & The story has its redundancies, and the young actors, not very experienced, are sometimes inexpressive. \textbf{Negative} \\
IV: & \underline{redundancies} and \underline{young actors} and \underline{experienced} and \underline{inexpressive} \\
OV: & \underline{Negative} \\
\bottomrule
\end{tabularx}
\caption{Examples of \emph{Input View} (IV) and \emph{Output View} (OV) in both tasks.}
\label{dualviewexample}
\end{table}

\section{Experiments}
\label{expsec}
This section first introduces experimental setup in Sec~\ref{setup}, and then presents main experiment results in Sec~\ref{mainresult}. Sec~\ref{ablation} conducts ablation study. In Sec~\ref{semi}, We compare \da and unlabeled data, present diversity analysis and a case study. 

\subsection{Experimental Setup}
\label{setup}
We conduct experiments on Sentence Classification tasks SST2~\cite{socher-etal-2013-recursive} and RT~\cite{pang-lee-2005-seeing} and Sequence Labeling tasks CoNLL03~\cite{tjong-kim-sang-de-meulder-2003-introduction} and Wikiann~\cite{pan-etal-2017-cross}. For each benchmark, we conduct shot-{10, 20, 50, 100} experiment. In Shot-$K$, we sample $K$ labeled instances for each output tag from the full training data. We repeatedly experiments 5 times and report the averaged micro-F1. The \textbf{Baseline} model is \emph{BERT-BASE} model only trained with few-shot training data $\gT$. Given the newly generated synthetic data $\gT_{LM}$, we train the same \emph{BERT-BASE} model using the same set of hyper-parameters. In sequence labeling tasks, we use rule-based data augmentation method \textbf{SDANER}~\cite{dai-adel-2020-analysis} and \textbf{MetaST}~\cite{wang2021meta}, a state-of-the-art self-training method, requiring additional \emph{unlabeled in-domain data}. For sentence classification tasks, rule-based \textbf{EDA}~\cite{wei-zou-2019-eda}, Back-Translation (\textbf{BackT.}) and bert-based \textbf{CBERT} methods are used. We adapt \textbf{LAMBADA}~\cite{notenoughpaper} as a PLM-based method for all tasks.

\paragraph{Implementation Details}
\da is built on the top of the T5-Large model~\cite{JMLR:v21:20-074}. \da requires Prompt Pre-training and fine-tuning with down-stream tasks. In both stages, we use Adafactor optimizer~\cite{shazeer2018adafactor} with learning rate 1e-3 and weight decay 1e-5 to train the \emph{Soft Prompt} parameters. For pre-training, we use the \emph{realnewslike} split in the T5 pre-training corpus C4 as the input. The pre-training batch size is 72 and we pre-train \da for 100k steps. We split the \emph{realnewslike} dataset into train and development split (i.e., 10000 pages). We will check the PPL on the development split every 5,000 steps. We save the model with lowest PPL. When fine-tuning on the few-shot data $\gT$, we set the batch size 32 and we train \da for 1,000 steps. We only upgrade the fine-tuning step to 5,000 on the shot-50 and shot-100 for Wikiann and CoNLL03. More experiment setup see Section~\ref{sec:appendix} in the Appendix.

\begin{figure*}[!htb]
\minipage{0.25\textwidth}
  \includegraphics[width=\linewidth]{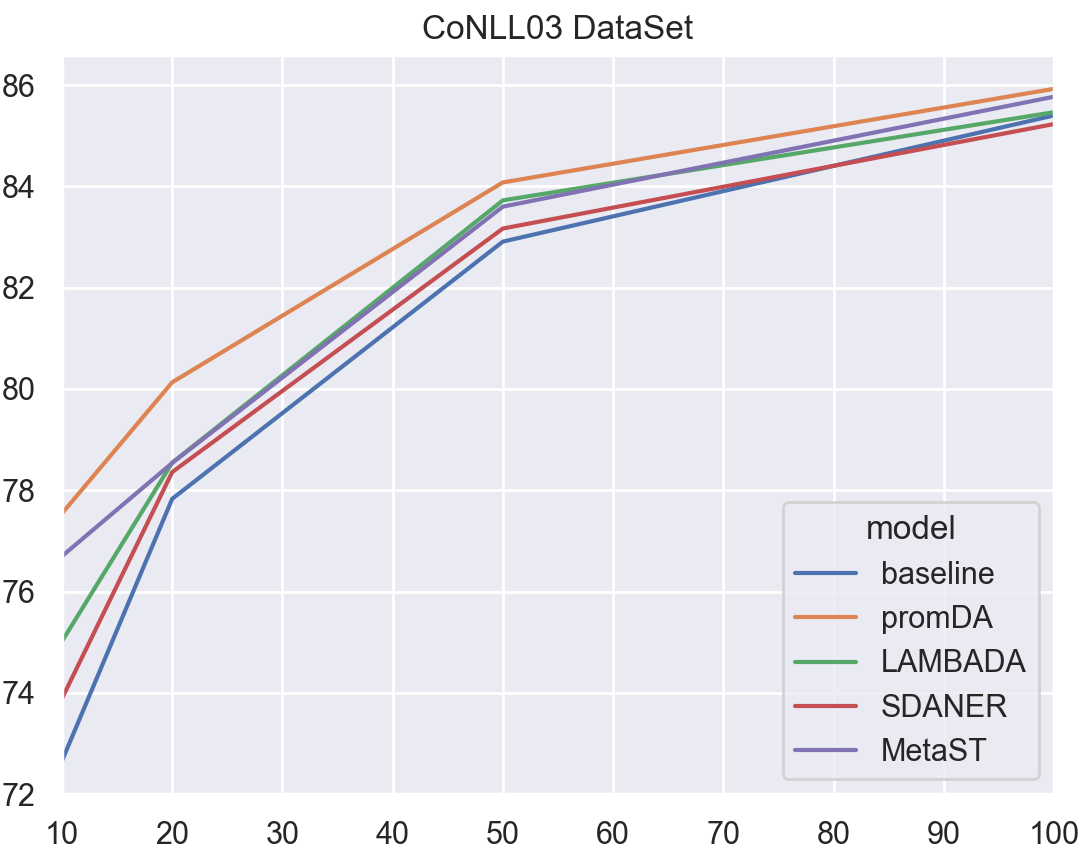}
\endminipage\hfill
\minipage{0.25\textwidth}
  \includegraphics[width=\linewidth]{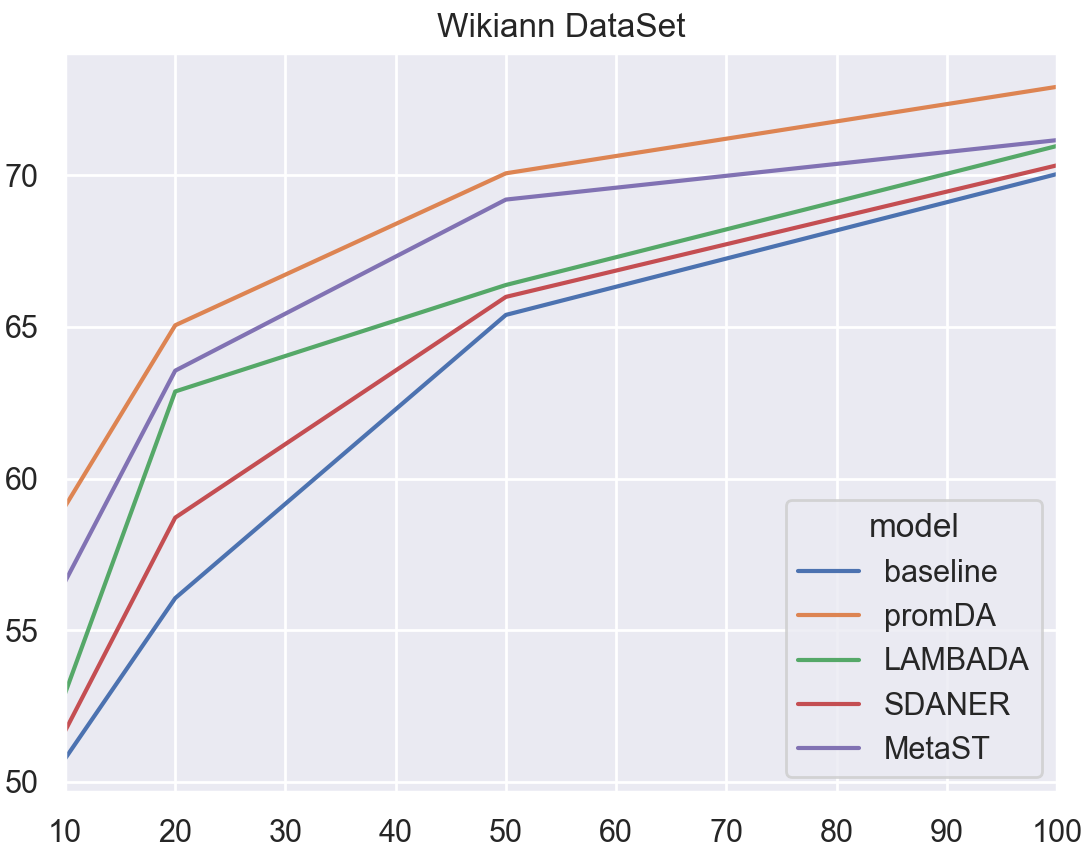}
\endminipage\hfill
\minipage{0.25\textwidth}%
  \includegraphics[width=\linewidth]{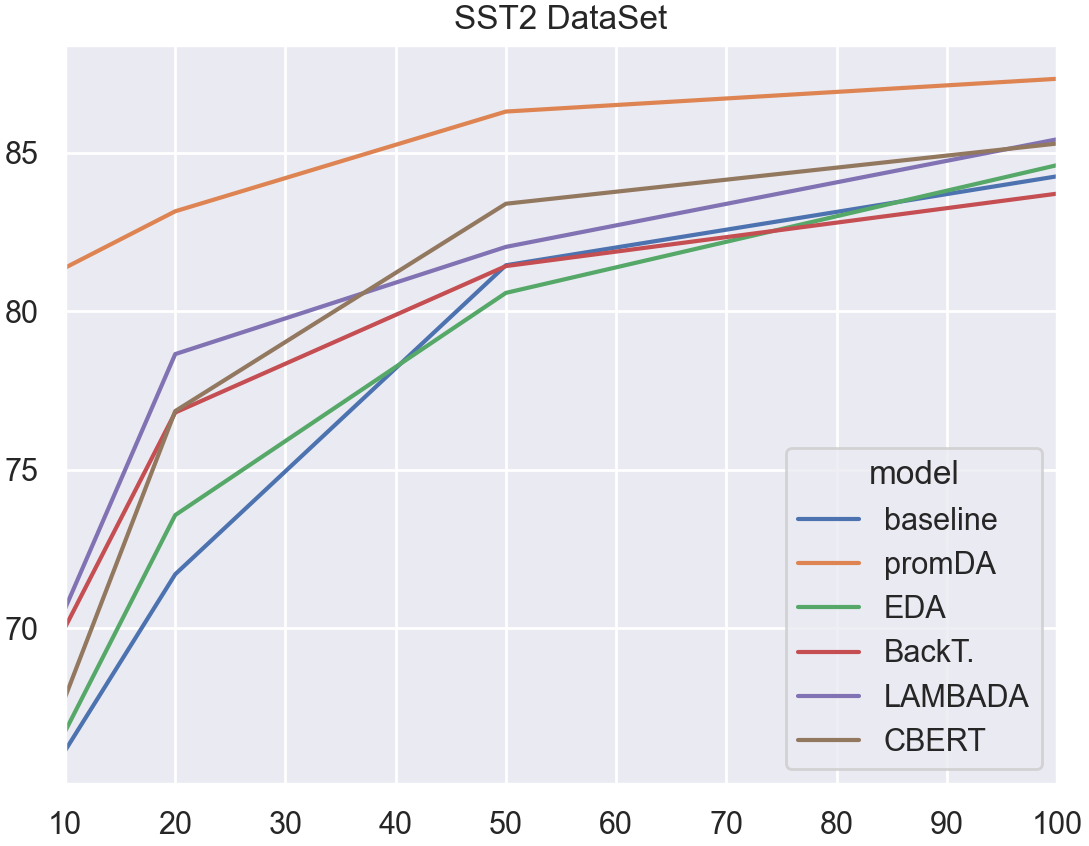}
\endminipage
\minipage{0.25\textwidth}%
  \includegraphics[width=\linewidth]{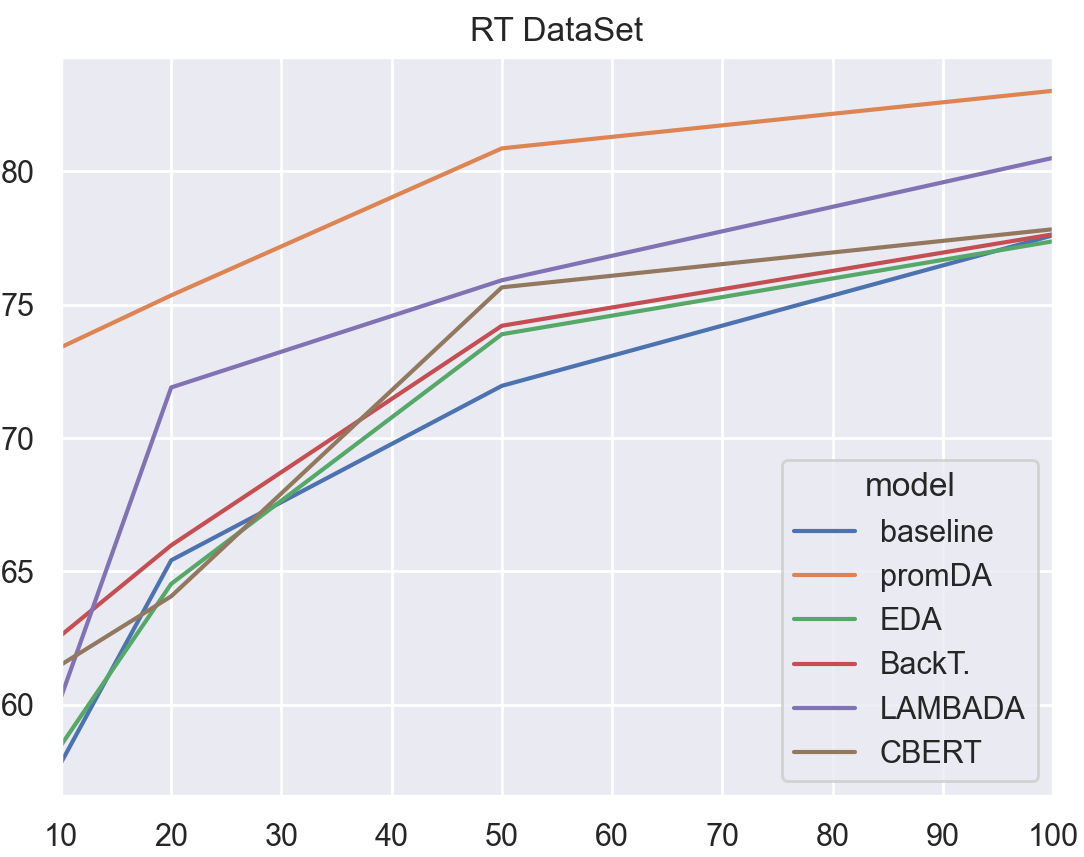}
\endminipage
\caption{Experiment results under the Shot-\{10, 20, 50, 100\} settings.}
\label{allsetting}
\end{figure*}

\subsection{Main Results}
\label{mainresult}

\paragraph{Sequence Labeling Tasks}
Table~\ref{seqlabelresult} summarizes the experiment results in shot-10 and shot-50. In both settings, the performance of NLU models trained with the synthetic data from \da are boosted up by a large margin (i.e., 4.8\% and 7.5\% for CoNLL03 and Wikiann, respectively).
\da also outperforms rule-based \textbf{SDANER} and fully fine-tuned PLM \textbf{LAMBADA} methods. In general, PLM-based approaches produce better synthetic data than \textbf{SDANER} does. Surprisingly, the NLU models supported by \da achieve slightly better performance than \textbf{MetaST} which uses \emph{unlabeled in-domain data}. This shows that \da could potentially reduce extra human effort in collecting \emph{unlabeled in-domain data} for the low-resource NLU tasks. Figure~\ref{allsetting} shows the performance in the shot-\{10, 20, 50, 100\} settings. The NLU models supported by \da consistently outperform other systems in all settings. Compared to Wikiann, the improvement margin in CoNLL03 is smaller. This could because the performance of CoNLL03 baseline is relatively high. 

\begin{table}[!ht]
\centering
\begin{tabularx}{0.48\textwidth}{cYYYY}
\bottomrule
DataSet         & \multicolumn{2}{c}{C03} & \multicolumn{2}{c}{Wiki} \\
\cmidrule{1-5}
Shot    & 10      & 50      & 10      & 50      \\
\bottomrule
Baseline & 72.7         & 82.9         & 50.8         & 65.4         \\
SDANER$^\spadesuit$ & 72.9 & 82.8 & 51.7 & 65.8 \\
LAMBADA & 75.0 & 83.7 & \underline{52.9} & \underline{66.4} \\
MetaST$^\clubsuit$   & 76.7         & 83.6         & 56.6         & 69.2 \\
\da     & \underline{77.5}         & \underline{84.1}         & \underline{58.3}         & \underline{70.1}         \\
\bottomrule
\end{tabularx}
\caption{Experiment Results of the Sequence Labeling Tasks. $^\clubsuit$ results taken from~\cite{wang2021meta}. $^\spadesuit$ we run ~\citet{dai-adel-2020-analysis}'s source code. C03 refers to CoNLL03 and Wiki refers to Wikiann. \underline{Underline} are the significant results compared to the \textbf{Baseline} model (paired student’s t-test, p < 0.05).}
\label{seqlabelresult}
\end{table}

\paragraph{Sentence Classification Tasks}
Table~\ref{senclsresult} shows the experiment results in shot-10 and shot-50. Similar to the results in the sequence labeling tasks, adding the synthetic data from \da significantly boosts up the performance of NLU models (more than 10\% in both benchmarks in shot-10). \da also outperforms various competitive methods, including \textbf{BackT.}, \textbf{CBERT} and \textbf{LAMBADA}. Although \textbf{LAMBADA} has higher level of flexibility and generates synthetic data from output tags, it only performs similar to \textbf{CBERT}. This could be because of the over-fitting issues when fine-tuning with small training data. Prompt-empowered \da successfully avoids this issue and produce high-quality synthetic data to support the NLU model training. Figure~\ref{allsetting} shows the performance in the shot-\{10, 20, 50, 100\} settings. NLU models supported by \da consistently outperform all other systems in all setups.

\begin{table}[!ht]
\centering
\begin{tabularx}{0.48\textwidth}{cYYYY}
\bottomrule
DataSet         & \multicolumn{2}{c}{SST2} & \multicolumn{2}{c}{RT} \\
\cmidrule{1-5}
Shot    & 10      & 50      & 10      & 50      \\
\bottomrule
Baseline & 66.1         & 81.5         & 57.8         & 72.0         \\
EDA$^\spadesuit$     & 66.7         & 80.4         & 58.5         & 73.9         \\
Back T. & 70.0 & 81.4 & 62.6 & 74.2 \\
CBERT$^\clubsuit$ & 67.8 & 83.4 & 61.5 & 75.3 \\
LAMBADA & 70.6 & 82.0 & 60.3 & 75.9 \\
\da     & \underline{81.4}         & \underline{86.3}         & \underline{73.4}         & \underline{80.9}         \\
\bottomrule
\end{tabularx}
\caption{Experiment Results of the Sentence Classification Tasks. $^\spadesuit$ we run ~\citet{wei-zou-2019-eda}'s source code. $^\clubsuit$ we run ~\citet{wu2019conditional}'s source code. \underline{Underline} are the significant results compared to the \textbf{Baseline} model (paired student’s t-test, p < 0.05).}
\label{senclsresult}
\end{table}


\paragraph{Discussion}
\textbf{LAMBADA} performs consistently worse than \da (e.g., more than 10\% F1 score gap in the SST2 and RT experiment). This is because fully fine-tuned PLMs can easily \emph{memorize} the limited labeled training data and produce similar synthetic data. In contrast, the prompt-based learning allows \da to maintain high generalization ability and provide new training signals to the NLU models. The results from \da are all statistical significant, compared to the \textbf{Baseline} model (paired student’s t-test, p < 0.05).

\subsection{Ablation Study}
\label{ablation}
We conduct ablation study for the components \emph{Prompt Pre-training}, \emph{Dual-View Data Augmentation} and \emph{Consistency Filtering} on the CoNLL03 and SST2 Benchmark under the shot-10 setting. 

\paragraph{Prompt Pre-Training} 
In \textbf{No PT}, we directly fine-tune two separated PLMs to learn the \emph{Input View} and \emph{Output View}. In \textbf{No PT Pre-Training}, we remove the Prompt Pre-training Task (\emph{Synonym Keywords to Sentence}). In \textbf{Full Pre-Training}, we apply the Prompt Pre-training Task to fine-tune the whole PLMs parameters. Finally, in \textbf{LM Adaptation}: we replace \da with solution in~\citet{lester2021power}. As shown in Table~\ref{ablationresult}, the fully fine-tuned PLMs (\textbf{No PT}) performs worse than our proposed \da method (4.6\% F1 score lower), showing the positive contribution of \emph{Soft Prompt} for low-resource NLU Data Augmentation. Further, removing PT Pre-training (\textbf{No PT Pre-Training}) or applying PT Pre-training to fine-tune all PLMs parameters (\textbf{Full Pre-Training}) also delegate the PT Pre-training performance by 3.1\% and 6.0\% F1 score, respectively, showing the importance of using PT Pre-training to learn a reasonable prompt initialization. Similarly, \textbf{LM Adaptation} also fine-tunes the whole PLMs and achieves similar performance as \textbf{Full Pre-Training}. It is recommended to directly train the prompt parameters. 

\begin{table}[!ht]
\centering
\begin{tabularx}{0.5\textwidth}{cYYY}
\bottomrule
DataSet         & C03 & SST2 & Ave. \\
\bottomrule
Few-shot NLU Baseline & 72.7 & 66.1 & 69.4 \\
\da & 77.5 & 81.4 & 79.5 \\
\specialrule{.4pt}{0pt}{0pt}
\emph{\footnotesize{Ablation for PT Pre-Training}} \\ [-0.6ex]
No PT & 75.2 & 74.5 & 74.9 \\
No PT Pre-Training & 74.0 & 78.2 & 76.1\\
Full Pre-Training & 75.0 & 72.0 & 73.5\\
LM Adaptation &  75.4 & 73.3 & 74.4 \\
\specialrule{.4pt}{0pt}{0pt}
\emph{\footnotesize{Ablation for Dual-View DA}} \\ [-0.6ex]
Output Only & 75.6 & 81.0 & 78.0\\
Input Only & 74.4 & 70.6 & 72.5\\
Single Prompt & 76.7 & 79.5 & 78.1 \\
\bottomrule
\end{tabularx}
\caption{Ablation Study for Prompt Pre-Training and Dual-View Data Augmentation for CoNLL03 and SST2 Benchmark under shot-10 settings.}
\label{ablationresult}
\end{table}

\paragraph{Dual-View Data Augmentation}
Next, we show the effect of Dual-View Data Augmentation in \da. \textbf{Input Only} and \textbf{Output Only} only generate synthetic data via the \emph{Input View} and \emph{Output view}, respectively. These two \emph{Single-View} models generate the same number of synthetic data as the \da does. As shown in Table~\ref{ablationresult}, 
the synthetic data from these two \emph{Single-View}  models successfully boost up the NLU model performance. However, their corresponding NLU models perform worse than the ones supported by \da. This shows that synthetic data from different views provide meaningful and different training signals to the NLU models. Interestingly, NLU models trained on the \emph{Output view}  perform better than the ones trained on the \emph{Input View}, indicating that output tags are more expressive signals to guide PLMs to generate high-quality synthetic data. Finally, instead of training two views on the separated prompt parameters, we train two views on the same prompt parameters in \textbf{Single Prompt}. The NLU models trained on \textbf{Single Prompt} synthetic data perform worse than the NLU models supported by \da, showing the importance of \emph{Prompt Ensemble} for \emph{Dual-View Data Augmentation}.

\begin{table}[!ht]
\centering
\begin{tabularx}{0.5\textwidth}{YcYYY}
\bottomrule
Setup &  w/o Filtering  & Iter-1 & Iter-2 & Iter-3 \\
\bottomrule
C03 & 72.0 & 76.7 & 77.6 & 77.5 \\
SST2 & 69.2 & 77.5 & 79.7 & 81.4 \\
\bottomrule
\end{tabularx}
\caption{Ablation Study For Iteration-based NLU Consistency Filtering.}
\label{ablationiter}
\end{table}

\paragraph{Consistency Filtering}
Finally, we examine the effect of \emph{Consistency Filtering} in \da. In table~\ref{ablationiter}, we show the NLU model performance without any filtering (\textbf{w/o Filtering}) and with $k$ iteration (\textbf{Iter-1}, \textbf{Iter-2} and \textbf{Iter-3}). The filtering has an important effect on the NLU performance. Without removing low-quality synthetic data, the performance gap almost disappears. The iteration filtering also has a positive effect on the NLU performance. In particular, in the SST2 Benchmark, the NLU model performance increases \textasciitilde 4\% F1 score after three iterations.

\begin{table}[!ht]
\centering
\begin{tabularx}{0.48\textwidth}{lYYYYY}
\bottomrule
Dataset & C03      & Wiki & SST2      & RT & $\Delta$   \\
\bottomrule
Baseline & 72.7 & 50.8 & 66.1 & 57.8 & - \\
w/ UID & 76.2 & 55.2 & 70.2 & 59.7 & +3.5 \\
w/ UND & 71.5 & 51.3 & 69.3 & 59.4 & +1.0 \\
w/ UGD & 64.6 & 44.8 & 66.4 & 58.7 & -3.2 \\
\midrule
\da & 77.5 & 58.3 & 81.4 & 73.4 & +10.8  \\
w/ UID & 80.0 & 61.7 & 83.0 & 73.9 & +12.8 \\
\bottomrule
\end{tabularx}
\caption{Experiment Results for \da and \emph{Unlabeled Data} under the shot-10 setting.}
\label{unlabeldatasource}
\end{table}

\begin{table*}[!ht]
\small
\centering
\begin{tabularx}{\linewidth}{rL}
\toprule
\multicolumn{2}{c}{{\bf Sequence Labeling}} \\
\midrule
GT: & \textcolor{red}{It quoted an} \textbf{[\textit{Org} Interior Ministry]} statement \textcolor{red}{as saying} \textbf{[\textit{Per} Shabir Ahmad Muhammad Jalil]} was executed in \textbf{[\textit{Loc} Mecca]}. \\
SDANER: & It , an \textbf{[\textit{Org} Interior Ministry]} statement as. \textbf{[\textit{Per} Mike Lloyd Muhammad Jalil]} was , in \textbf{[\textit{Loc} Berlin]} \\
LAMBADA: & \textcolor{red}{It quoted an} \textbf{\textcolor{blue}{\textbf{[\textit{Misc} German]}}} newspaper \textcolor{red}{as saying} \textbf{\textcolor{blue}{\textbf{[\textit{Per} Klaus Kinkel]}}} was \textbf{\textcolor{blue}{killed in \textbf{[\textit{Loc} Berlin]} plane crash.}} \\
\da: & The \textbf{\textcolor{blue}{\textbf{[\textit{Org} AP]}}} \textcolor{red}{quoted} the head of  \textbf{\textcolor{blue}{\textbf{[\textit{Org} the Back of New Zealand]}}} \textcolor{red}{as saying} \textbf{\textcolor{blue}{``\textbf{[\textit{Per}Noel Davis]} said the bank would cut its ties with \textbf{[\textit{Org} Bancorp]} and \textbf{[\textit{Per} Peter Johnson]} would be the next to take over.''}} \\
\midrule
GT: & \textbf{[\textit{Loc} U.S.]} embassy officials in \textbf{[\textit{Loc} Colombo]} \textcolor{red}{were not immediately available to comment on the report}. \\
SDANER: & \textbf{[\textit{Loc} Mahala]} embassy jail in \textbf{[\textit{Loc} China]} were not immediately available to v on the board. \\
LAMBADA: & \textbf{[\textit{Loc} U.S.]} embassy officials in \textbf{[\textit{Loc} Saudi Arabia]} \textcolor{red}{were not immediately available to comment on the report.} \\
\da: & The \textbf{[\textit{Loc} United States]} Embassy in \textbf{[\textit{Loc} Germany]} \textbf{\textcolor{blue}{has not commented on the latest developments}} \\
\midrule
GT: & \textcolor{red}{\textbf{[\textit{Org} All Fishermen 's Association]} secretary} \textbf{[\textit{Per} N.J. Bose]} \textcolor{red}{said the strike would continue indefinitely and the fishermen would block road and rail traffic if their demands were not met.} \\
LAMBADA: & \textcolor{red}{\textbf{[\textit{Org} All Fishermen's Association]} secretary} \textcolor{blue}{\textbf{[\textit{Per} N. Fairbrother]}} \textcolor{red}{said the strike would continue indefinitely and the fishermen would block road and rail traffic if their demands were not met.} \\
\da: & \textcolor{blue}{\textbf{[\textit{Org} FIA]}} secretary \textcolor{blue}{\textbf{[\textit{Per} David Davidson]}} said the 
\textcolor{blue}{\textbf{[\textit{Org} World Bank]}} had offered them about 60 million euros. \\
\midrule
GT: & \textbf{[\textit{Org} Araneta Coliseum]}, \textbf{[\textit{Loc} Quezon City]}; \textbf{[\textit{Org} University of Akron]}, \textbf{[\textit{Loc} Akron]}  \\
\da: & \textcolor{blue}{\textbf{[\textit{Org} Memorial Theatre]}, \textbf{[\textit{Loc} Houston, Texas]}; \textbf{[\textit{Org} University of Louisville]}, \textbf{[\textit{Loc} Louisville]}}\\
\midrule
\multicolumn{2}{c}{{\bf Sentence Classification}} \\
\midrule
GT1: & \textcolor{red}{The story has its redundancies}, and the young actors, not very experienced, are sometimes inexpressive. \textbf{Negative} \\
GT2: & Full frontal, which opens today nationwide, could almost be classified as a movie-industry satire, but it lacks the generous inclusiveness that is \textcolor{red}{the genre 's definitive, if disingenuous}, feature. \textbf{Negative} \\
LAMBADA: & \textcolor{red}{The story has its redundancies}, the bullying that are \textcolor{red}{the genre's definitive, if disingenuous}, episodes. \textbf{Negative} \\
\da: & Despite their experience, the young actors are sometimes inexpressive because of the redundancies in the story, \textbf{\textcolor{blue}{which may have caused them to feel rushed and out of place}}. \textbf{Negative} \\
\midrule
GT: & I could just feel \textcolor{red}{the screenwriter} at every moment ``tap , tap , tap , tap , tapping away'' on this screenplay. \textbf{Negative} \\
\da: & \textbf{\textcolor{blue}{Mr. Mcsweeney}}, \textcolor{red}{the screenwriter}, has done a masterful job with \textbf{\textcolor{blue}{a dynamic plot, full of suspense, wit, and humor}}. \textbf{Positive} \\
\midrule
GT: & \textcolor{red}{The saigon of 1952} is an uneasy mix of sensual delights and simmering violence , and the quiet american brings us right into the center of that world .	\textbf{Positive} \\
LAMBADA: & many of the ladies in blue are very much in love with the \textcolor{red}{saigon of 1952}, and many of them are very much in love with the country. \textbf{Positive} \\
\da: & The \textbf{\textcolor{blue}{saigon of the '70s}} is the antithesis of that, with a sardonic tone and well conceived plot that builds toward a great end.  \textbf{Positive} \\
\bottomrule
\end{tabularx}
\caption{Generated synthetic data from our proposed \da and other baseline methods. Text chunks in \textcolor{red}{Red} are duplicated with the few-shot training data. Text chunks in \textcolor{blue}{\textbf{Blue}} are the novel words/phrases.}
\label{casetable}
\end{table*}

\subsection{Discussion}

\paragraph{\da with T5-Base}
We verify whether \da could work with different pre-trained language models. We replace the T5-Large model with the T5-base model. The new \da can also improve the few-shot baseline models by a large margin. On the SST2 shot-10 setup, the NLU model is improved from 66.1 to 76.3 F1 score, which also beats other models presented in Table~\ref{senclsresult}.

\paragraph{\da in the high-resource setting}
To show the advantages of \da in the high-resource setting, We replace the few-shot training data with the full training data. We find that \da can still improve the baseline model performance. In SST2, after adding syntactic data, the NLU performance is improved from 90.8 to 92.3 F1 score.

\paragraph{Improvement Margin Difference}
As shown in Table~\ref{seqlabelresult} and~\ref{senclsresult}, the improvement margins in the sentence classification tasks (i.e., more than 15\% F1 score) are generally larger than the ones in the sequence labelling tasks (i.e., less than 10\% F1 score). This could because \emph{i)} the sequence labelling task is a more fine-grained and knowledge-intensive task than the sentence classification task; \emph{ii)} the synthetic data for the sequence labelling tasks includes entity type and boundary, which is more challenging for PLMs to generate, in particular for low-resource settings, compared to the sentence classification task.

\paragraph{\da and Unlabeled Data}
\label{semi}
The above experiments are based on the assumption that no \emph{unlabeled data} is available. In this section, we explore the connection between \da and \emph{unlabeled data}. To incorporate \emph{unlabeled data} into our NLU models, we apply the classic \emph{self-training} framework~\cite{1053799} to the NLU models. Specifically, for each unlabeled instance, we use the NLU models to label it and record the output tags and corresponding likelihood score. The low likelihood score means predictions with less confidence. We rank all unlabeled instances based on the likelihood score and remove instances at the bottom 20\%. Table~\ref{unlabeldatasource} shows the experiment result of four benchmarks under the shot-10 setting.

\paragraph{The Effect of Unlabeled Data Domain}
We design three settings: \emph{Unlabeled In-domain Data} (\textbf{UID}), \emph{Unlabeled Near-domain Data} (\textbf{UND}) and \emph{Unlabeled General-domain Data} (\textbf{UGD}) where the unlabeled data come from \emph{exactly same}, \emph{similar} and \emph{general-purpose} domains. We exchange the training data between CoNLL03 and Wikiann, and between SST2 and RT to simulate \emph{similar} domains. We randomly sample sentences from PLM pre-training corpus to simulate the \emph{general-purpose} domain. We note that unlabeled data domain has a great impact of the \emph{self-training} performance. Even a slight domain shift (i.e., \textbf{UND}) delegates the NLU performance by 2.5\%. 
The performance of NLU models trained with unlabeled data from general-purpose corpus are even 3.2\% lower than the NLU baseline models only trained with few-shot labeled data $\gT$. Both sequence labeling tasks and sentence classification tasks follow this trend, but sequence labeling tasks is more sensitive to the unlabeled data domain. Extra human effort is still required, for semi-supervised learning, to select suitable domains to collect unlabeled data. 

\paragraph{Combining \emph{Unlabeled In-domain Data} with \da}
We apply the above \emph{self-training} algorithm to the final NLU models (\da) supported by \da with \emph{unlabeled in-domain data}. The resulting NLU models are further improved, on average, by 2.0\% (\textbf{w/ UID} in the last row). More sophisticated semi-supervised learning algorithms may introduce more improvement. This shows that \emph{a)} synthetic data from \da and \emph{unlabeled in-domain data} provide different information to the NLU models; \emph{b)} \da successfully extracts the embedded knowledge in the PLMs and presents them in the generated synthetic data.

\paragraph{Diversity Analysis}
\label{diversity}
In Table~\ref{diverse}, we show the diversity of the generated synthetic data from \da and other baseline models. We sample 10 new synthetic data from each training instance. We use \textbf{Novel Mention} (number of entity mentions or keywords not appearing in the training data) and Self-BLEU score~\cite{zhu2018texygen} to measure the diversity. In general, simple generative data augmentation approaches (i.e, \textbf{BackT.} and \textbf{CBERT}) can easily produce Novel Mentions, but their generated synthetic data lacks diversity (relatively low self-BLEU score). The prompt-based learning helps \da to produce the most diverse synthetic data with the most Novel Mentions in both benchmarks. Due to the over-fitting issues, \textbf{LAMBADA} produces synthetic data that are less or equal diverse than other baseline approaches. Interestingly, the NLU models trained on these synthetic data achieve the second best performance. This could because \textbf{LAMBADA} coherently generate the whole synthetic sentences, while others reply on the random and/or heuristic rules.

\begin{table}[!ht]
\begin{tabularx}{0.5\textwidth}{cYYY}
\bottomrule
Model  & NM$\uparrow$    & Self-B$\downarrow$ & F1$\uparrow$  \\
\specialrule{.4pt}{0pt}{0pt}
\emph{\footnotesize{CoNLL03}} \\ [-0.6ex]
SDANER & 141.4  & 0.770  & 72.9   \\
LAMBADA  & 107.6   & 0.761  & 75.0   \\
\da   & \textbf{351} & \textbf{0.259}  & \textbf{77.5}   \\
\specialrule{.4pt}{0pt}{0pt}
\emph{\footnotesize{SST2}} \\ [-0.6ex]
EDA    & 59.6    & 0.889  & 66.7   \\
BackT. & 101.8 & 0.826  & 70.0   \\
CBERT  & 127    & 0.900  & 67.8   \\
LAMBADA  & 51.8  & 0.926  & 70.6   \\
\da   & \textbf{276} & \textbf{0.578}  & \textbf{81.4}   \\
\bottomrule
\end{tabularx}
\caption{Diversity Analysis for the generated synthetic data in CoNLL03 and SST2 under the shot-10 settings. NM refers to Novel Mentions.}
\label{diverse}
\end{table}

\paragraph{Synthetic Data Case Study}
\label{case}
Table~\ref{casetable} shows representative examples generated by our proposed \da and methods. In the Sequence Labelling example, the rule-based \textbf{SDANER} shuffles the original word order and creates low-quality text. The \textbf{LAMBADA} model generates a new synthetic instance by modifying three text spans in the original training instance (e.g., changing ``statement'' to ``newspaper''). In contrast, Our \da method generates a completely new and reasonable event in a bank, as well as correct and novel geographical locations in the generated synthetic data. 
Similarly, in the sentence classification tasks, \textbf{LAMBADA} naively combines text chunks from two training instances in the second example. \da mentions some keywords in the training data, but adds more information into the output. In another example, \da comments on a screenwriter (not appearing in the training data) with a sequence of coherent words. Finally, \da successfully moves the topic from the film ``The Saigon of 1952'' to the Saigon in 70s. In summary, \da can extract the embedded real-world knowledge from the PLMs and introduces these knowledge into a relatively long sentence in a fluent way.

\section{Conclusion and Future Work}
In this paper, we present the first prompt-based pre-trained language model \da for low-resource NLU data augmentation. Experiments on four benchmarks show the effectiveness of our proposed \da method. In the future, we plan to expand \da to other NLP tasks, including question answering, machine reading comprehension and text generation tasks. 

\section*{Acknowledgement}
We thank anonymous reviewers for their insightful suggestions to improve this paper. Yufei Wang, Can Xu, Qingfeng Sun, Huang Hu, Chongyang Tao, Xiubo Geng and Daxin Jiang are supported by Microsoft Software Technology Center at Asia (STCA). Yufei Wang also receives a MQ Research Excellence Scholarship and a CSIRO’s DATA61 Top-up Scholarship.

\bibliography{anthology,custom}
\bibliographystyle{acl_natbib}

\clearpage
\appendix
\section{Experiment Details}
\label{sec:appendix}


\subsection{Implementation Details for NLU model}
We use \emph{BERT-BASE} as our NLU models. The \textbf{Baseline} model is only trained with the few-shot training data $\gT$. Given the newly generated synthetic data, we will train the same NLU model with the same set of hyper-parameters. The only difference between the two NLU models is the training data. To train the \emph{BERT-BASE} model, we use the Adam optimizer to train the model with learning rate 5e-5 and weight decay 5e-6. We train all NLU models with 4,000 steps and check the validation performance every 400 steps. We use batch size 8. 

\subsection{Implementation Details for Compared Models}
\textbf{EDA}~\footnote{\url{https://github.com/jasonwei20/eda_nlp}} and \textbf{SDANER}~\footnote{\url{https://github.com/boschresearch/data-augmentation-coling2020}} are rule-based data augmentation methods. They modify the available training instances via simple rules, including word order shuffle, synonym replace, etc. Since they have released their source code on GitHub, we directly run their source code, without any modification, for our experiments.  \textbf{BackT.} first translates the input sentence in language A to language B, and then translates back to language A, which may create new linguistic expressions in the back-translated sentences. We directly use the \emph{M2M100} model~\cite{fan2021beyond}, without any fine-tuning, to translate the sentence from English to French and backwards. \textbf{CBERT}~\cite{wu2019conditional} uses BERT model to replace words in the input sentences. Compared to EDA, the decision is made based on the context information, which should be more accurate. We use the suggested parameters and code released by the authors~\footnote{\url{https://github.com/1024er/cbert_aug}}. We Implement the \textbf{LAMBADA} model based on its original paper~\cite{notenoughpaper}. The only difference is that, to allow a fair comparison with our proposed \da method, we replace its PLMs (i.e., GPT2) with T5-Large model. For \textbf{LM adaptation}, we follow the fine-tuning configuration in its original paper~\cite{lester2021power}.

\subsection{Trainable Parameters}
\da adds 5 trainable vectors at each encoder layer of the frozen T5-Large model. The total trainable parameters in \da is 2 * 5 * 24 * 1024 = 245760 (2 for two sets of \emph{Soft Prompt} for \emph{Input View} and \emph{Output View}). This parameter scale is very closed to the \textbf{LM Adaptation} approach which has 2 * 100 * 1024 = 204800 trainable parameters. 

\subsection{Dual-View Data Augmentation}
As shown in Alg.~\ref{alg:data_diverse}, we train $M_{LM}$ using few-shot data $\gT$. We then feed the keywords in $\gT$ to the \emph{Input View} and the output label sequence to the \emph{Output View}. We duplicate each instance in $\gT$ 40 times before feeding them into \da for generation. We use the standard nucleus sampling~\cite{Holtzman2020The} with top\_p = 0.9. For each input sequence, we sample 5 output sequences. Finally, we duplicate each instance in $\gT$ 100 times, then combine them with $\gT^r_{LM}$. For iteration-based NLU Consistency Filtering, we find that iterating 3 times is a powerful filtering strategy.

\subsection{Computing Infrastructure and Running Time}
We use Nvidia A100 and V100 for our experiment. A single A100 or V100 is capable to handle the T5-Large model. In general, it takes around 6-8 hours to generate synthetic data for few-shot training data $\gT$ with 300 - 400 instances.

\subsection{Evaluation Metrics}
We report averaged Micro-F1 (short for micro-averaged F1 score), which assesses the quality of multi-label binary problems by measuring the F1-score of the aggregated contributions of all classes, for the 5 times for each of our experiment. We also conduct statistical test using the paired t-student test between the baseline model results and \da method. We use the implementation of scipy~\footnote{\url{https://docs.scipy.org/doc/scipy/reference/generated/scipy.stats.ttest_rel.html}} to calculate p values. All of \da result are statistical significant (p < 0.05).

\section{Dataset}

\subsection{Evaluation Source}
As for the evaluation benchmarks, the CoNLL03 and Wikiann dataset are from the repository of MetaST~\cite{wang2021meta}~\footnote{\url{https://github.com/microsoft/MetaST}}.  CoNLL03 and
Wikiann are public benchmarks for Named
Entity Recognition. CoNLL03 is a collection of news wire articles from the Reuters Corpus with manual annotations, whereas Wikiann comprises of extractions from Wikipedia. The SST2 (Stanford Sentiment Tree-bank) and RT (a movie review corpus from Rotten Tomatoes) dataset are from the repository of CBERT~\cite{wu2019conditional}~\footnote{\url{https://github.com/1024er/cbert_aug}}.

\subsection{Training data for different Few-shot Settings}
Table~\ref{number} shows the number of training data in different few-shot settings.

\begin{table}[!ht]
\begin{tabularx}{0.5\textwidth}{cYYYY}
\bottomrule
Shot                     & 10 & 20 & 50  & 100 \\ \cmidrule{1-5}
CoNLL03                  & 40 & 80 & 200 & 400 \\ 
Wikiann                  & 30 & 60 & 150 & 300 \\ 
SST2                     & 20 & 40 & 100 & 200 \\ 
RT & 20 & 40 & 100 & 200 \\
\bottomrule
\end{tabularx}
\caption{The new of training data instances for each benchmark under different shot-k settings.}
\label{number}
\end{table}

\section{Experiment Analysis}

\subsection{Shot-20 and Shot-100 Results}
Table~\ref{seqlabelresultshot20shot100} and~\ref{senclsresultshot20shot100} show the concrete performance of \da and other baseline models under the shot-20 and shot-100 settings. It is interesting to note that \textbf{F.LMs} often outperforms other baseline models in the shot-100 setting. This could because \textbf{F.LMs} avoids over-fitting and starts to learn to generate novel mentions when the few-shot training data becomes larger. 

\begin{table}[!ht]
\centering
\begin{tabularx}{0.5\textwidth}{YYYYY}
\bottomrule
DataSet         & \multicolumn{2}{c}{C03} & \multicolumn{2}{c}{Wiki} \\
\cmidrule{1-5}
Shot    & 20      & 100      & 20      & 100      \\
\bottomrule
Baseline & 77.8         & 85.4         & 56.1         & 70.0         \\
SDANER$^\spadesuit$ & 78.4 & 85.2 & \underline{58.7} & 70.3 \\
F.LMs & 78.6 & 85.5 & \underline{62.9} & \underline{71.0} \\
MetaST$^\clubsuit$   & 78.5         & 85.8         & 63.6         & 71.2 \\
\da     & \underline{80.1}         & 85.9         & \underline{65.1}         & \underline{72.9}         \\
\bottomrule
\end{tabularx}
\caption{Experiment Results of the Sequence Labelling Tasks. $^\clubsuit$ results taken from~\cite{wang2021meta}. $^\spadesuit$ we run ~\citet{dai-adel-2020-analysis}'s source code. C03 refers to CoNLL03 and Wiki refers to Wikiann. \underline{Underline} are the significant results compared to the \textbf{Baseline} model (paired student’s t-test, p < 0.05).}
\label{seqlabelresultshot20shot100}
\end{table}

\begin{table}[!ht]
\centering
\begin{tabularx}{0.5\textwidth}{YYYYY}
\bottomrule
DataSet         & \multicolumn{2}{c}{SST2} & \multicolumn{2}{c}{RT} \\
\cmidrule{1-5}
Shot    & 20      & 100      & 20      & 100      \\
\bottomrule
Baseline & 71.7         & 84.3         & 65.4         & 77.6         \\
EDA$^\spadesuit$     & 73.6         & 84.6         & 64.5         & 77.4         \\
BackT. & \underline{76.8} & 83.7 & 66.0 & 77.6 \\
CBERT$^\clubsuit$ & \underline{76.9} & 85.3 & 64.1 & 77.8 \\
F.LMs & \underline{78.7} & 85.4 & \underline{71.9} & 80.5 \\
\da     & \underline{83.2}         & \underline{87.3}         & \underline{75.4}         & \underline{83.0}         \\
\bottomrule
\end{tabularx}
\caption{Experiment Results of the Sentence Classification Tasks. $^\spadesuit$ we run ~\citet{wei-zou-2019-eda}'s source code. $^\clubsuit$ we run ~\citet{wu2019conditional}'s source code. \underline{Underline} are the significant results compared to the \textbf{Baseline} model (paired student’s t-test, p < 0.05).}
\label{senclsresultshot20shot100}
\end{table}

\subsection{Unlabeled Data Domain}
In Sec~\ref{semi}, we analysis three types of unlabeled data: \emph{Unlabeled In-domain Data} (\textbf{UID}), \emph{Unlabeled Near-domain Data} (\textbf{UND}) and \emph{Unlabeled General-domain Data} (\textbf{UGD}). We will give details on how these three types of unlabeled data are constructed. The \emph{Unlabeled In-domain Data} are the training instances in the original full training data but not included in the current few-shot training set $\gT$. When used as unlabeled data, we ignore their supervised labels. Those training instances are from the exactly same source and therefore, they are guaranteed to be in the \emph{same} domain. We exchange the training data between CoNLL03 and Wikiann, and between SST2 and RT as \emph{Unlabeled Near-domain Data} to simulate \emph{similar} domains. This is because that \emph{1)} both CoNLL03 and Wikiann have \emph{Person}, \emph{Organization} and \emph{Location}; \emph{2)} both SST2 and RT are reviews in daily life. Finally, we randomly sample 10,000 sentences from the T5 pre-training corpus to simulate the \emph{general-purpose} domain.

\subsection{Diversity Metrics}
In Sec~\ref{diversity}, we use two metrics, \textbf{Novel Mention} and \textbf{Self-Bleu}, to measure the diversity of generated synthetic data. \textbf{Novel Mention} is defined as the entity mention or keywords that do not appearing  in the training data. For the sequence labelling tasks, we directly extract the named entity mentions from each instance as the \emph{Mentions}. For the sentence classification tasks, we extract top-3 keywords from the input sentence using the unsupervised keyword extract \emph{Rake}~\cite{rose2010automatic} as the \emph{Mentions}. The higher \textbf{Novel Mention} is, the better. \textbf{Self-Bleu} evaluates how one sentence resembles the rest in a generated collection. The lower \textbf{Self-Bleu} is, the better.

\end{document}